\documentclass{article}

\usepackage{microtype}
\usepackage{graphicx}
\usepackage{caption}
\usepackage{placeins}
\usepackage{booktabs} 
\usepackage{multicol}
\usepackage{multirow}
\usepackage{wrapfig}
\usepackage{microtype}

\usepackage{hyperref}

\usepackage{bm}
\newcommand{\matr}[1]{\bm{#1}}
\newcommand*{\tran}{^\top}

\usepackage{comment}

\usepackage[accepted]{icml2023_dp4ml}

\usepackage{amsmath}
\usepackage{amssymb}
\usepackage{mathtools}
\usepackage{amsthm}
\usepackage{empheq}
\usepackage{lipsum}

\usepackage[capitalize,noabbrev]{cleveref}

\theoremstyle{plain}
\newtheorem{theorem}{Theorem}[section]
\newtheorem{proposition}[theorem]{Proposition}
\newtheorem{lemma}[theorem]{Lemma}
\newtheorem{corollary}[theorem]{Corollary}
\theoremstyle{definition}
\newtheorem{definition}[theorem]{Definition}

\theoremstyle{remark}

\DeclareMathOperator{\Tr}{\mathrm{Tr}}
\DeclareMathOperator{\St}{\mathrm{St}}

\usepackage[textsize=tiny]{todonotes}

\icmltitlerunning{Duality in Multi-view Restricted Kernel Machines}

\begin{document}

\twocolumn[
	\icmltitle{Duality in Multi-View Restricted Kernel Machines}



	\icmlsetsymbol{equal}{*}

	\begin{icmlauthorlist}
		\icmlauthor{Sonny Achten}{equal,yyy}
		\icmlauthor{Arun Pandey}{equal,yyy}
		\icmlauthor{Hannes De Meulemeester}{yyy}
		\icmlauthor{Bart De Moor}{yyy}
		\icmlauthor{Johan A.K. Suykens}{yyy}


	\end{icmlauthorlist}

	\icmlaffiliation{yyy}{KU Leuven, Department of Electrical Engineering (ESAT), STADIUS Center for
Dynamical Systems, Signal Processing and Data Analytics}

	\icmlcorrespondingauthor{Sonny Achten}{sonny.achten@kuleuven.be}

	\icmlkeywords{Machine Learning, ICML}

	\vskip 0.3in
]



\printAffiliationsAndNotice{\icmlEqualContribution} 

\begin{abstract}
	We propose a unifying setting
	that combines existing restricted kernel machine methods
	into a single primal-dual multi-view framework for kernel principal component analysis in both supervised and unsupervised settings.
	We derive the primal and dual representations of the framework and relate different training and inference algorithms from a theoretical perspective. We show how to achieve full equivalence in primal and dual formulations by rescaling primal variables.
	Finally, we experimentally validate the equivalence and provide insight into the relationships between different methods on a number of time series data sets by recursively forecasting unseen test data and visualizing the learned features. 
\end{abstract}

\section{Introduction}

Kernel methods have seen continued success over the years in many applications due to their provable guarantees and strong theoretical foundations \cite{vapnik95, Scholkopf2001, suykens_least_2002, gp_machine_learning_Rasmussen06}. Kernel methods allow for a complicated non-linear problem in the input dimension $\mathcal{X}$ to be solved in a high-dimensional Reproducing Kernel Hilbert Space $\mathcal{H}$ by using a feature map $\phi(\cdot): \mathcal{X} \rightarrow \mathcal{H}$ to transform the input data. Additionally, if the optimization problem can be expressed by only using inner products, then the problem can be restated in terms of a kernel function $k(\bm{x}, \bm{y})=  \langle \phi(\bm{x}), \phi(\bm{y})\rangle_{\mathcal{H}}  $. This trick is commonly referred to as the kernel trick and is particularly useful when one wishes to work with very-high dimensional (possibly infinite) feature maps.
This concept led to numerous popular methods such as kernel principal component analysis~\citep{kpca_97,kpca}, kernel Fisher discriminant analysis \citep{kfda}, support vector machines \cite{vapnik95} and least-squares support vector machines~\citep{suykens_least_2002}.
While kernel methods have shown excellent performance and generalization capabilities, they tend to fall behind when it comes to large-scale problems due to their memory and computational complexity. Additionally, it can be difficult to change their architecture to allow for hierarchical representation learning, which is one of the most powerful capabilities of neural networks.
Recently, Restricted Kernel Machines (RKM), were proposed which connect least-squares support vector machines and
kernel principal component analysis (kernel PCA) with Restricted Boltzmann machines \cite{suykensdeep2017}.
RKMs extend the primal and dual model representations present in least-squares support vector machines, from shallow to deep architectures by introducing the dual variables as hidden features through conjugate feature duality.
This provides a framework of kernel methods represented by visible and hidden units as in Restricted Boltzmann Machines \cite{hinton2005kind}.

Although it is well known that there is a primal-dual equivalence in principal component analysis in feature space and kernel principal component analysis, as an eigendecomposition of the covariance matrix and kernel matrix respectively \cite{Scholkopf2001, kpca_lssvm}, little attention has been paid to the proper scaling of the obtained components. Furthermore, optimization algorithms on the Stiefel manifold have been proposed as an alternative solver to replace the eigendecompositions, but they do not give a clear interpretation of the eigenvalues, which is however important for out-of-sample extensions and missing view imputation in a multi-view setting, especially when one does not want to train a pre-image map.

\noindent \textbf{Contributions.}
(i) We derive a primal and dual setting from a single objective function, and theoretically show the equivalence between the two settings by obtaining a scaling for the primal variables.
(ii) Furthermore, we show that the optimization on the Stiefel manifold yields, in general, a non-diagonal matrix, which is a rotational transformation of a diagonal eigenvalue matrix.
(iii) Additionally, we study known algorithms in their elementary form, and propose minor but significant modifications to achieve full equivalence between different training algorithms.
(iv) We validate the theoretical arguments on standard time-series datasets and provide code for reproducibility.

\begin{figure*}[t]
	\centering
	\def\svgwidth{0.8\linewidth}
	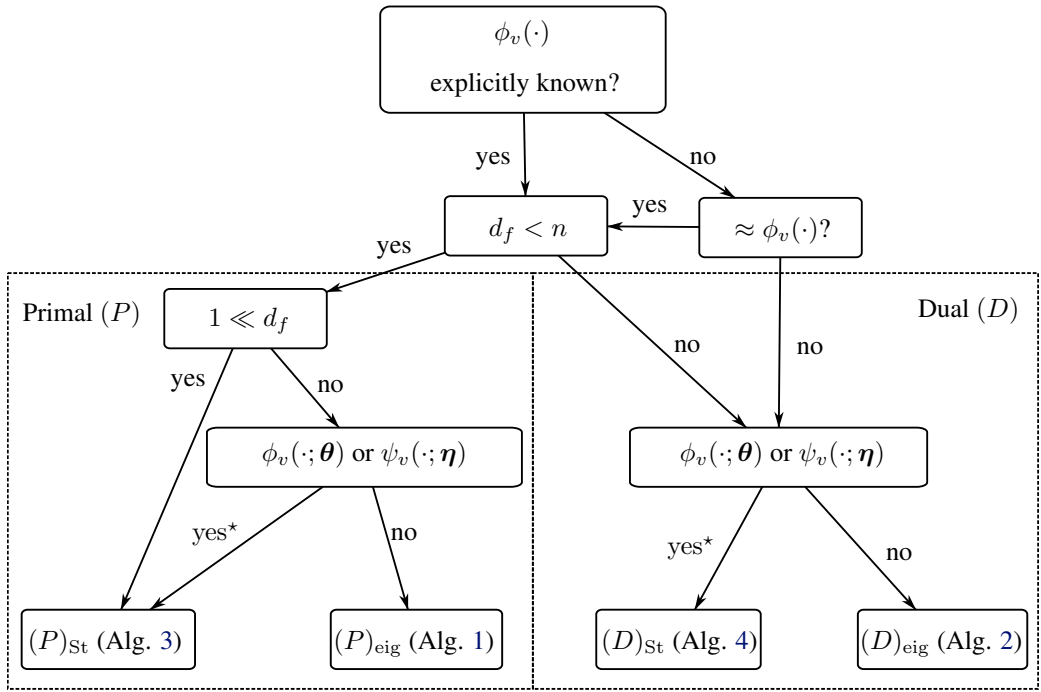
	\caption{Flowchart showing which training algorithm to choose depending on the number of datapoints $n$,  feature space dimension $d_f = \sum_{v=1}^{V}d_{f,v}$ with $V$ views with corresponding feature maps $\phi_{v}(\cdot)$ / pre-image maps $\psi_{v}(\cdot)$ are explicitly known or not (or can be approximated $\approx \phi_{v}(\cdot)$?), and whether they have trainable parameters $\bm{\theta}$, $\bm{\eta}$. $^\star$ denotes that the training objective should be augmented with an additional loss term (e.g., reconstruction error).}
	\label{fig:schematic_flowchart}
\end{figure*}

\section{Related Works}

\noindent \textbf{Restricted Kernel Machines.}
The restricted kernel machines framework ~\cite{suykensdeep2017} provides a general setting for classification, regression and unsupervised learning by employing kernel PCA. For each of these settings, the corresponding LS-SVM formulation \cite{suykens_least_2002} is given an interpretation in terms of visible and hidden variables, achieved by the Legendre-Fenchel duality for a quadratic function \cite{Rockafellar_duality}. This framework has been extended for many new settings and has been used as the basis for multiple novel models.
RKMs have been shown to be excellent at learning disentangled representations in the form of features~\cite{WinantLatentSpace,TONIN2021661}.
The disentangled representation learning capability has shown to be useful in unsupervised out-of-distribution detection \cite{ToninOOD}.
A deep variant of kernel PCA has been developed in the RKM framework, where it is shown that the multiple levels allow for learning more efficient disentangled representations than standard kernel PCA while still being explainable \cite{tonin2023deep}. The standard classification method has been extended to the multi-view setting by using tensor-based representations \cite{HOUTHUYS202154}.
Generative models have been created using the RKM framework, such that the latent space can be sampled to create novel data from the learned distribution. Additionally, the inherent disentangled features allows the latent space to be traversed along the learned principal components to generate samples with certain characteristics \cite{joachim, GENRKM, strkm}.
Finally, a model for time-series was proposed by introducing autoregression amongst the latent variables of an RKM model \cite{rrkm_esann}.
This method was further extended to create a general multi-view kernel PCA method to learn and forecast time-series data \cite{pandey_multi-view_2023}.

\noindent \textbf{Energy-Based Methods.}
Energy-based models use a so-called energy function  to model the joint-probability distribution over visible and latent variables of a system. The training process consists of finding a configuration of the learnable parameters on the energy function with the lowest total energy. The advantage of these types of models is that the energy function does not require normalization as opposed to fully probabilistic methods. An excellent overview of energy-based learning can be found in \citep{tutorial_energy_based}.
One of the most well known energy-based models is the Restricted Boltzmann Machine~\citep{salakhutdinov_restricted} which has a representation of the energy function based on visible and hidden units with no hidden to hidden connections. This layer-based approach with bipartite connections between neighboring layers allows for efficient learning and for deeper architectures to be created \citep{dbm}.

\section{Theoretical Framework}
This Section builds the theoretical framework in a primal-dual setting for multi-view kernel principal component analysis, which forms the basis for subsequent sections. Note that the framework is set up for an arbitrary number of views, and that this therefore also holds for unsupervised learning with one view.

The data consists of $n$ datapoints in $\mathcal{X}=\mathcal{X}_1 \times \ldots \times \mathcal{X}_V$, with $\mathcal{X}_v = \{\bm{x}_{v,i}\}_{i=1}^{n}$, and $\bm{x}_{v,i}\in \mathbb{R}^{d_v}$. $V\in \mathbb{N}$ is the number of \emph{views} (also called modalities). 

Conventionally, a view is data, or a representation thereof, from a single source of data or modality related to a problem. Multi-view methods use these different views of the data together to solve a specific task. For example, using neural networks it is common for different views to be mapped to a joint space to create a richer data representation  \cite{survey_datafusion}.
Considering kernel methods, multi-view learning typically attempts to combine the kernels of the different views together \cite{gonen2011multiple, mkl_multiview_spec_cluster}.

Although the framework is essentially unsupervised, one can cast it into a supervised problem by using one of the views as labels or targets. The inference scheme within such a setting is shown in~\cref{sec:prediction}. All proofs are given in Appendix \ref{app:proofs}.

We start from a super-objective $J_{\text{init}}$, for both training and prediction. We generalize \citet[Equation (3.24)]{suykensdeep2017} to multiple views, with the difference that we now add regularization terms for the feature maps $\phi_v(\cdot)$. We consider the following constrained minimization problem:
\begin{multline}
	\min_{\bm{U}_{v}, \  \phi_v(\bm{x}_{v,i}), \  \bm{e}_{v,i}} J_{\text{init}} = \frac{1}{2}\sum_{v=1}^V\Tr(\bm{U}_v\tran \bm{U}_v)
	\\
	-\frac{V}{2}\sum_{v=1}^V\sum_{i=1}^{n}\bm{e}_{v,i}\tran\bm{\Gamma}^{-1}\bm{e}_{v,i}
	+\frac{1}{2}\sum_{v=1}^V\sum_{i=1}^{n}
	\Vert \phi_v(\bm{x}_{v,i})\Vert_2^2
	\\
	\text{s.t. } \quad \bm{e}_{v,i}=\bm{U}_v\tran\phi_v(\bm{x}_{v,i}) \quad \forall v,i,
	\label{eq:initial_optimization_problem}
\end{multline}
where $\bm{\Gamma}$ is a symmetric positive definite hyperparameter matrix, $\phi_v(\cdot): \mathcal{X}_v \rightarrow \mathcal{H}_v$  are feature mappings for the different data views, and $\bm{U}_v$ are transformation matrices. For further use, we define $\Phi_v = [\phi_v(\bm{x}_{v,1}), \dots ,\phi_v(\bm{x}_{v,n})]\tran \in \mathbb{R}^{n\times \dim(\mathcal{H}_v)}$, $\bm{C}_{vw}=\bm{\Phi}_v\tran \bm{\Phi}_w$.
Without loss of generality, we assume that the feature maps are zero centered. In primal setting, one can easily achieve this by subtracting the means of the columns of $\Phi_v$. When using implicit feature maps in the dual setting, one can do this by left and right multiplying the kernel matrix with the centering matrix $\bm{M}_c=\mathbb{I}-\frac{1}{n}\bm{1}\bm{1}\tran$ \citet[Appendix]{pandey_multi-view_2023}.
We introduce dual variables $\bm{h}_i$ by means of Fenchel-Young inequality on the score variables $\bm{e}_{v, i}\in\mathbb{R}^s$: 
\begin{equation}
	\frac{V}{2}\bm{e}\tran_{v,i}\bm{\Gamma}^{-1}\bm{e}_{v,i}+\frac{1}{2V}\bm{h}_i\tran\bm{\Gamma}\bm{h}_i\geq \bm{e}_{v,i}\tran\bm{h}_i \quad \forall v,i.  \label{eq:fy-inequality}
\end{equation}
Conjugating the same feature representation $\bm{h}_i$ with different score variables $\bm{e}_{v,i}$ gives rise to a duality gap. However, the goal which we achieve is that the new objective will now obtain a \emph{coupling} between the views $v$. The above inequality could be verified by writing it in quadratic form:
\begin{equation}
	\frac{1}{2}
	\begin{bmatrix}
		\bm{e}^{\top}_{v,i} & \bm{h}^{\top}_{i} \\
	\end{bmatrix}
	\begin{bmatrix}
		V\bm{\Gamma}^{-1} & -\mathbb{I}_s          \\
		-\mathbb{I}_s     & \frac{1}{V}\bm{\Gamma}
	\end{bmatrix}
	\begin{bmatrix}
		\bm{e}_{v,i} \\
		\bm{h}_{i}
	\end{bmatrix}
	\geq
	0, \quad \forall v,i,
	\label{eq:quadratic_form_fenchel_young}
\end{equation}

With $\mathbb{I}_s$ the $s$-dimensional identity matrix. \Cref{eq:quadratic_form_fenchel_young} follows immediately from the Schur complement form.
It states that for a matrix $\bm{Q} = \left[\begin{smallmatrix}
			\bm{A} & \bm{B} \\ \bm{B}^{\top} & \bm{C} \end{smallmatrix}\right],$
one has $\bm{Q} \succeq 0$ if and only if $\bm{A} \succ 0$ and the
Schur complement $\bm{C} - \bm{B}^{\top} \bm{A}^{-1} \bm{B} \succeq 0$
\cite{Boyd:2004:CO:993483}.

\begin{definition}[Primal and Dual]
	We further refer to $\bm{U}_v$ as the \emph{primal variables} and $\bm{h}_{i}$ as the \emph{dual variables}, and likewise for extensions to their (concatenated) matrix representations. The \emph{primal setting (P)} is the setting in which all dual variables are eliminated from the equations and the \emph{dual setting (D)} is the setting in which all primal variables are eliminated from the equations.
\end{definition}

Typically, feature maps in the dual setting only appear in inner products. Further, when substituting \eqref{eq:fy-inequality} in \eqref{eq:initial_optimization_problem} and eliminating the score
variables $\bm{e}_{v, i}$, we obtain a primal-dual minimization problem with objective $J_{\text{pr,d}} \geq J_{\text{init}} $ :
\begin{multline}
	\min_{\bm{U}_v,\bm{h}_i,\phi_v(\bm{x}_{v,i})} \ J_{\text{pr,d}} \triangleq \frac{1}{2}\sum_{v=1}^V\Tr(\bm{U}_v\tran \bm{U}_v)+\frac{1}{2}\sum_{i=1}^{n}\bm{h}_{i}\tran\bm{\Gamma}\bm{h}_{i}\\
	-\sum_{v=1}^V\sum_{i=1}^n\phi_v(\bm{x}_{v,i})\tran \bm{U}_v \bm{h}_i
	+\frac{1}{2}\sum_{v=1}^{V}\sum_{i=1}^{n} \Vert \phi_v(\bm{x}_{v,i})\Vert_2^2.
	\label{eq: primal-dual optimization problem}
\end{multline}
The first order conditions for optimality of this optimization problem are given by the stationarity conditions:
\begin{empheq}[left=\empheqlbrace]{align}
	\dfrac{\partial J_{\text{pr,d}}}{\partial \bm{U}_v}=0 \Rightarrow \ & \bm{U}_v=\sum_{i=1}^n \phi_v(\bm{x}_{v,i})\bm{h}_i\tran, \label{eq:SC1}\\
	\dfrac{\partial J_{\text{pr,d}}}{\partial \bm{h}_i}=0 \Rightarrow \ & \bm{h}_i = \bm{\Gamma}^{-1}\sum_{v=1}^V\bm{U}_v\tran\phi_v(\bm{x}_{v,i}),\label{eq:SC2}\\
	\dfrac{\partial J_{pr,d}}{\partial \phi_v(\bm{x}_{v,i})}
	= 0 \Rightarrow \ & \phi_v(\bm{x}_{v,i}) =
	\bm{U}_{v}\bm{h}_{i}.\label{eq:SC3}
\end{empheq}
Note that problem \eqref{eq: primal-dual optimization problem} is generally non-convex.
Whether it has a solution depends on the hyperparameters $\bm{\Gamma}$. In the next subsections, we show how to determine $\bm{\Gamma}$ automatically, while formulating a solution for either the primal or dual variables and showing how they are related.

\subsection{Primal formulation}

We continue with solving \eqref{eq: primal-dual optimization problem} w.r.t. the primal variables $\bm{U_v}$. By eliminating the dual variables from \eqref{eq: primal-dual optimization problem} using \eqref{eq:SC2}, one can obtain:
\begin{multline*}
	\min_{{\bm{U}}_v, \ \phi_v(\bm{x}_{v,i})} -\frac{1}{2}\sum_{v,w=1}^V \Tr (\bm{\Gamma}^{-\frac{1}{2}}{\bm{U}}_v\tran \bm{C}_{vw}{\bm{U}}_w\bm{\Gamma}^{-\frac{1}{2}})\\+\frac{1}{2}\sum_{v=1}^V\Tr(\bm{U}_v\tran\bm{U}_v)+\frac{1}{2}\sum_{v=1}^V\Tr(\bm{C}_{vv}).
\end{multline*}
The above expression is the inspiration for the next Lemma, which helps us to reformulate the problem without the need of defining $\bm{\Gamma}$, by relating it to the Lagrange multipliers $\bm{Z}$ of the following constrained optimization problem:
\begin{lemma}\label{lemma:primal_equivalence}
	Given that $\bm{\Gamma}=(\bm{Z}+\bm{Z}\tran)/2=\tilde{\bm{Z}}$ and that $\bm{U}_v=\tilde{\bm{U}}_v\tilde{\bm{Z}}^{1/2}$,
	the solution to the primal minimization problem:
	\begin{multline}\label{eq:primal_optimization_problem}
		\min_{\tilde{\bm{U}}_v, \ \phi_v(\bm{x}_{v,i})}  J_{\text{pr}} \triangleq
		-\frac{1}{2}\sum_{v,w=1}^V \Tr (\tilde{\bm{U}}_v\tran \bm{C}_{vw}\tilde{\bm{U}}_w)\\+\frac{1}{2}\sum_{v=1}^V\Tr(\bm{C}_{vv})\\
		\text{s.t.} \ \tilde{\bm{U}}\tran \tilde{\bm{U}} =
		\left[\begin{array}{ccc}
				\tilde{\bm{U}}_1\tran & \cdots & \tilde{\bm{U}}_V\tran
			\end{array}\right]
		\left[\begin{array}{c}
				\tilde{\bm{U}}_1 \\
				\vdots           \\
				\tilde{\bm{U}}_V
			\end{array}\right]
		=\mathbb{I}_s,
	\end{multline}
	satisfies the same first order conditions for optimality w.r.t. primal variables $\bm{U}_v$ as \eqref{eq: primal-dual optimization problem}.

\end{lemma}

The introduced constraint does not only yield equivalence in first order optimality conditions, but it also keeps the objective bounded. Further, from the KKT conditions of \eqref{eq:primal_optimization_problem} (see \eqref{eq:primal_KKT_1} and \eqref{eq:primal_KKT_3}), and since $\bm{\Gamma}=\tilde{\bm{Z}}$, one can derive:
\begin{equation}
	\bm{\Gamma}=\sum_{v,w}^V\tilde{\bm{U}}_v\tran\bm{C}_{vw}\tilde{\bm{U}}_w=\tilde{\bm{U}}\tran\bm{C}\tilde{\bm{U}},
\end{equation}
with cross-covariance matrix
\begin{equation}
	\label{eq:C_mat}
	\bm{C}=\left[{\begin{smallmatrix}
					\bm{C}_{11} & \cdots & \bm{C}_{1V} \\
					\vdots        & \ddots & \vdots        \\
					\bm{C}_{V1} & \cdots & \bm{C}_{VV}
				\end{smallmatrix}}\right].
\end{equation}

For a given feature map and training data, \eqref{eq:primal_optimization_problem} is guaranteed to have a minimizer. Indeed, it is a minimization of a concave objective over a \emph{compact set} \cite{Boyd:2004:CO:993483}. Note that \eqref{eq:primal_optimization_problem} can be solved by a gradient-based algorithm.
\begin{proposition} \label{proposition:primal_span}
	Given a symmetric and positive definite matrix $\bm{B}\in\mathbb{R}^{m\times m}$
	with real eigenvalues $\lambda_1 \geq \dots \geq \lambda_n > \lambda_{n+1} \geq \dots \geq  \lambda_{n} \geq 0$ and corresponding eigenvectors $\bm{v}_1, \dots, \bm{v}_{n}$ ; then $\bm{A}\in\mathbb{R}^{m\times n}$ is a minimizer of
	\begin{equation*}
		\min_{\bm{A}}-\frac{1}{2}\Tr(\bm{A}\tran \bm{B}\bm{A})
		+\frac{1}{2}\Tr(\bm{B}) \quad
		\text{s.t.} \ \bm{A}\tran \bm{A}=\mathbb{I}_n, \label{eq:proposition_objective}
	\end{equation*}
	if and only if ${\bm{A}}^\top {\bm{A}}=\mathbb{I}_n \land \text{range}({\bm{A}})=\text{span}(\bm{v}_1, \dots, \bm{v}_n)$.
\end{proposition}

Given Lemma \ref{lemma:primal_equivalence} and Proposition \ref{proposition:primal_span}, we arrive at the following property for the solution in primal variables $\bm{U}_v$:
\begin{corollary}\label{cor:Primal_rotations}
	Given a feature map and datapoints, the solution to the eigendecomposition problem:
	\begin{equation}
		\bm{C}
		\tilde{\bm{U}}
		=  \tilde{\bm{U}}\bm{\Lambda},
	\end{equation}
	with $\bm{U}=\tilde{\bm{U}}\bm{\Lambda}^{1/2}=\tilde{\bm{U}}(\tilde{\bm{U}}\tran\bm{C}\tilde{\bm{U}})^{1/2}$ is a globally optimal solution of \eqref{eq:primal_optimization_problem}. In this case, $\bm{\Gamma}=\bm{\Lambda}$ are diagonal. Also, any orthonormal projection  $\tilde{\bm{U}}^\prime=\tilde{\bm{U}}\bm{O}\tran$ gives another optimal solution:
	\begin{equation*}
		\bm{C}
		\tilde{\bm{U}}^\prime
		=  \tilde{\bm{U}}^\prime\bm{\Gamma}^\prime,
	\end{equation*}
	with $\bm{U}^\prime=\tilde{\bm{U}}^\prime\bm{\Gamma}^{\prime1/2}$ and non-diagonal $\bm{\Gamma}^{\prime}=\bm{O}\bm{\Lambda}\bm{O}\tran$. Furthermore, any $\bm{U}^\prime$ satisfies the stationarity conditions of \eqref{eq: primal-dual optimization problem} w.r.t. the primal variables.
\end{corollary}
One can now see that the primal setting in case of a single view results in a principal component analysis in
the feature space, or some orthonormal projection. The rescaling of the principal components is the result of stationarity condition \eqref{eq:SC3} and is important when one wants to use the trained model to infer a view from the other views for a new datapoint (see further). For multi-view, we arrive at a principal component analysis where the feature vector is a concatenation of the feature vectors of the individual views.

\subsection{Dual formulation}

We now proceed with similar formulations, but in a dual setting. Let's
first introduce Gram matrices $\bm{K}_{v}=\bm{\Phi}_v
	\bm{\Phi}_v\tran$, $\forall v$, with its elements $\bm{K}_{v,ij} =
	\phi(\bm{x}_{v,i})\tran\phi(\bm{x}_{v,j})$.\footnote{We will use the term Gram matrix and kernel
	matrices interchangeably, for reasons explained in~\cref{sec:training_schemes}.} We then denote $\bm{H} = [ \bm{h}_1 , \ldots , \bm{h}_n ]^\top \in \mathbb{R}^{n\times s}$ and eliminate the primal variables from \eqref{eq: primal-dual optimization problem} using \eqref{eq:SC1}:
\begin{multline*}
	\min_{{\bm{H}}, \ \phi_v(\bm{x}_{v,i})} -\frac{1}{2}\sum_{v=1}^V \Tr (\bm{H}\tran \bm{K}_{v}{\bm{H}})\\+\frac{1}{2}\Tr(\bm{H}\tran\bm{H})+\frac{1}{2}\sum_{v=1}^V\Tr(\bm{K}_{v}),
\end{multline*}
after which we introduce a similar constrained optimization formulation in the dual setting:

\begin{lemma}\label{lemma:dual_equivalence}
	Given that $\bm{\Gamma}=(\bm{Z}+\bm{Z}\tran)/2=\tilde{\bm{Z}}$,
	the solution to the primal minimization problem:
	\begin{multline}\label{eq:dual_optimization_problem}
		\min_{\bm{H}, \ \phi_v(\bm{x}_{v,i})}\quad J_{\text{d}} \triangleq
		-\frac{1}{2}\sum_{v=1}^V\Tr(\bm{H}\tran \bm{K}_{v}\bm{H})
		\\+\frac{1}{2}\sum_{v=1}^V\Tr(\bm{K}_{v})
		\quad \text{s.t.} \ \bm{H}\tran \bm{H}=\mathbb{I}_s,
	\end{multline}
	satisfies the same first order conditions for optimality, with Lagrange multipliers $\bm{Z}$, w.r.t. dual variables $H$ as \eqref{eq: primal-dual optimization problem}.
\end{lemma}

From the first KKT condition of \eqref{eq:dual_optimization_problem} (see also \eqref{eq:dual_KKT_1}), one can derive an expression for $\bm{\Gamma}$:
\begin{equation}
	\bm{\Gamma}=\bm{H}\tran \big[\sum_{v=1}^V{\bm{K}}_v \big]\bm{H},
	\label{eq:kpca}
\end{equation}
and the second KKT condition (see also \eqref{eq:dual_KKT_3}) indicates that the dual variables $\bm{h}_i$ are orthonormal. By reinvoking Lemma \ref{lemma:primal_equivalence}, we can make the following claims regarding optimality:

\begin{corollary}
	Given a feature map and datapoints, the solution to the eigendecomposition problem:
	\begin{equation}
		\bm{K}
		{\bm{H}}
		=  {\bm{H}}\bm{\Lambda},
	\end{equation}
	is a globally optimal solution of \eqref{eq:dual_optimization_problem}. In this case, $\bm{\Gamma}=\bm{\Lambda}$ are diagonal. Also, any orthonormal projection  ${\bm{H}}^\prime={\bm{H}}\bm{O}\tran$ gives another optimal solution:
	\begin{equation*}
		\bm{K}
		{\bm{H}}^\prime
		=  {\bm{H}}^\prime\bm{\Gamma}^\prime,
	\end{equation*}
	with non-diagonal $\bm{\Gamma}^{\prime}=\bm{O}\bm{\Lambda}\bm{O}\tran$.
	Furthermore, any $\bm{H}^\prime$ satisfies the stationarity conditions of
	\eqref{eq: primal-dual optimization problem} w.r.t. the dual variables.

	\label{corr:kpca_dual}
\end{corollary}

In this dual setting, the formulation results in a kernel principal component analysis, or some orthonormal projection. The different views are incorporated by summing their kernel functions together.

\section{Training and Inference schemes}

\subsection{Training schemes}\label{sec:training_schemes}

\noindent \textbf{Algorithms.} The theoretical framework of the preceding section gives rise to different training algorithms. Not only can one choose to train a model in a primal or dual setting, but one can also decide to use an eigendecomposition or a constrained optimization algorithm. First, note that the feasible set for the constraints in problems \eqref{eq:primal_optimization_problem} and \eqref{eq:dual_optimization_problem} is the Stiefel manifold $\text{St}(m,n)=\{\matr{A} \in \mathbb{R}^{m \times n} \, | \, {\matr{A}}^{\top} \matr{A}=\mathbb{I}_{n}\}$ with $\bm{A}=\tilde{\bm{U}}$ and $\bm{A}=\bm{H}$ respectively. Therefore, one can use the Cayley Adam optimizer \cite{li2019} to solve \eqref{eq:primal_optimization_problem} or \eqref{eq:dual_optimization_problem}. For training in primal setting, one can employ  \cref{alg:training_eig_P} or \cref{alg:training_stiefel_P}. These algorithms yield the same solution up to some orthonormal transformation. Optionally, one can rotate the solution obtained by the latter algorithm to achieve full equivalence (see Corollary \ref{cor:Primal_rotations}). Analogously, one can employ \cref{alg:training_eig_D} or \cref{alg:training_stiefel_D}, and optionally rotate the latter, to train in a dual setting. After training, one can always use equations \eqref{eq:SC1} or \eqref{eq:SC2} to get the primal variables from the dual or vice versa, i.e., when the feature maps are explicitly known.

\begin{table*}[t]
	\centering
	\caption{Overview of different training algorithms in primal or dual setting, and with eigendecomposition or constrained optimization on the Stiefel manifold.}
	\label{tab:algorithms}
	\begin{tabular}{ccc}
		                                 & \textbf{Primal} & \textbf{Dual} \\
		\hline
		\begin{minipage}{0.01\textwidth}\rotatebox{90}{\small Eigendecomposition}\end{minipage}
		                                 &
		\begin{minipage}{0.42\textwidth}
			\begin{algorithm}[H]
				\caption{}
				\label{alg:training_eig_P}
				\begin{algorithmic}[1]
					\begin{small}
						\STATE {\bfseries Input:} $\Phi_v$ for all training points and views
						\STATE {\bfseries Output:} $\bm{U}, \bm{\Gamma}$
						\STATE center feature maps: \\ $\Phi_{v} \gets  \Phi_v - \bm{1}_n\sum_i\phi_v(\bm{x}_{v,i})\tran/n$
						\STATE construct cross-covariance matrix $\bm{C}$ \eqref{eq:C_mat}
						\STATE $\tilde{\bm{U}}, \bm{\Lambda} \gets \text{eig}(\bm{C})$
						\STATE ${\bm{U}} \gets \tilde{\bm{U}}\bm{\Lambda}^{1/2}, \ \ \bm{\Gamma} \gets \bm{\Lambda}$
					\end{small}
				\end{algorithmic}
			\end{algorithm}
		\end{minipage}
		                                 &
		\begin{minipage}{0.48\textwidth}
			\begin{algorithm}[H]
				\caption{}
				\label{alg:training_eig_D}
				\begin{algorithmic}[1]
					\begin{small}
						\STATE {\bfseries Input:} $\Phi_v$; or kernel functions and features $\bm{X}_v$ for all training points and views
						\STATE {\bfseries Output:} $\bm{H}, \bm{\Gamma}$
						\STATE construct kernel/Gram matrix $\bm{K}_v$ for all views
						\STATE center kernel matrices and sum: $\bm{K} \gets \sum_v\bm{M}_c\bm{K}_{v}\bm{M}_c$
						\STATE ${\bm{H}}, \bm{\Lambda} \gets \text{eig}(\bm{K})$
						\STATE $\bm{\Gamma} \gets \bm{\Lambda}$
					\end{small}
				\end{algorithmic}
			\end{algorithm}
		\end{minipage}                                    \\
		\begin{minipage}{0.01\textwidth}\rotatebox{90}{\small Stiefel}\end{minipage}
		                                 &
		\begin{minipage}{0.42\textwidth}
			\begin{algorithm}[H]
				\caption{}
				\label{alg:training_stiefel_P}
				\begin{algorithmic}[1]
					\begin{small}
						\STATE {\bfseries Input:} $\Phi_v$ for all training points and views
						\STATE {\bfseries Output:} $\bm{U}, \bm{\Gamma}$
						\STATE center feature maps: \\$\Phi_{v} \gets  \Phi_v - \bm{1}_n\sum_i\phi_v(\bm{x}_{v,i})\tran/n$
							\STATE construct cross-covariance matrix $\bm{C}$ \eqref{eq:C_mat}
							\STATE solve \eqref{eq:primal_optimization_problem}:  $\tilde{\bm{U}^{\prime}}\gets \text{CayleyAdam}(\bm{C})$
							\STATE $\bm{\Gamma}^{\prime} \gets \tilde{\bm{U}}^{\prime\top} \bm{C} \tilde{\bm{U}^{\prime}}$
							\IF{rotate solution}
							\STATE ${\bm{O}}, \bm{\Lambda} \gets \text{eig}(\bm{\Gamma}^{\prime})$
							\STATE  $\tilde{\bm{U}} \gets \tilde{\bm{U}^{\prime}}\bm{O}, \ \ \bm{\Gamma} \gets \bm{\Lambda}$
							\ELSE
							\STATE  $\tilde{\bm{U}} \gets \tilde{\bm{U}^{\prime}}, \ \ \bm{\Gamma}  \gets \bm{\Gamma}^{\prime}$
							\ENDIF
							\STATE ${\bm{U}} \gets \tilde{\bm{U}}\bm{\Gamma}^{1/2}$
					\end{small}
				\end{algorithmic}
			\end{algorithm}
			\vspace{0.001in}
		\end{minipage} &
		\begin{minipage}{0.48\textwidth}
			\begin{algorithm}[H]
				\caption{}
				\label{alg:training_stiefel_D}
				\begin{algorithmic}[1]
					\begin{small}
						\STATE {\bfseries Input:} $\Phi_v$; or kernel functions and features $\bm{X}_v$ for all training points and views
						\STATE {\bfseries Output:} $\bm{H}, \bm{\Gamma}$
						\STATE construct kernel/Gram matrix $\bm{K}_v$ for all views
						\STATE center kernel matrices and sum: \\$\bm{K} \gets \sum_v\bm{M}_c\bm{K}_{v}\bm{M}_c$
							\STATE solve \eqref{eq:dual_optimization_problem}: ${\bm{H}^{\prime}} \gets \text{CayleyAdam}(\bm{K})$
							\STATE $\bm{\Gamma}^{\prime}=\bm{H}^{\prime\top}\bm{K}\bm{H}^{\prime}$
							\IF{rotate solution}
							\STATE ${\bm{O}}, \bm{\Lambda} \gets \text{eig}(\bm{\Gamma}^{\prime})$
							\STATE $ {\bm{H}} \gets {\bm{H}^{\prime}}\bm{O}, \ \ \bm{\Gamma} \gets \bm{\Lambda}$
							\ELSE
							\STATE ${\bm{H}} \gets {\bm{H}^{\prime}}, \ \ \bm{\Gamma}  \gets \bm{\Gamma}^{\prime}$
						\ENDIF
					\end{small}
				\end{algorithmic}
			\end{algorithm}
			\vspace{0.001in}
		\end{minipage}                                    \\
		\hline
	\end{tabular}
\end{table*}

\noindent \textbf{Explicit or implicit feature maps.}
The feature maps $\phi_v(\cdot): \mathbb{R}^{d_v} \rightarrow \mathbb{R}^{d_{f,v}}$ maps the data view's input space $\mathbb{R}^{d_v}$ to a high-dimensional feature space $\mathbb{R}^{d_{f,v}}$. However, instead of explicitly defining these feature maps, one can apply the kernel trick:
${K}_{v,ij} = k_{v}(\bm{x}_{v,i}, \bm{x}_{v,j}) = \langle
	\phi_v(\bm{x}_{v,i}), \phi_v(\bm{x}_{v,j}) \rangle_{\mathcal{H}_v}$, where
$k_v(\cdot, \cdot): \mathbb{R}^{d_v} \times \mathbb{R}^{d_v} \mapsto \mathbb{R}$ is
the kernel function corresponding to a canonical feature map $\phi_v(\cdot)$ satisfying
the Mercer theorem.

When an explicit feature map is not known but a kernel function is used, one can only use the dual setting. Alternatively, one can use techniques such as Random Fourier Features \cite{random_fourier_features} or the Nyström Method \cite{nystrom_method} to approximate the canonical feature map of a kernel function.
In case a feature map (or approximation) is known, one can choose either the primal or dual setting. If this feature map is parametric: $\phi_v(\cdot;\bm{\theta})$, one can jointly train its parameters $\bm{\theta}$ using Algorithm \ref{alg:training_stiefel_P} or \ref{alg:training_stiefel_D} when augmenting the objective function with some reconstruction loss term \cite{GENRKM, strkm}.

\noindent \textbf{Schematic Overview.}
Although all training algorithms yield the same solutions, they scale differently w.r.t. the number of datapoints and feature space dimensions in terms of computation cost. Therefore, also considering the aforementioned, Figure \ref{fig:schematic_flowchart} provides a tool to decide which algorithm to use in which case.

\subsection{Inference Schemes}\label{sec:prediction}
For inference, one can use the framework to infer missing views of unseen datapoints.
To do this, the following problems need to be tackled.

\noindent \textbf{Inference equations}.
Similar to \citet{pandey_multi-view_2023}, we can derive inference equations from the stationarity conditions \eqref{eq:SC1} to \eqref{eq:SC3}. More precisely, one can obtain an expression for $\phi_v(\bm{x}_{v,i})$ with all other $\phi_w(\bm{x}_{w,i})$ or $\bm{k}_v(\bm{x}_{v,i})$ with all other $\bm{k}_w(\bm{x}_{w,i})$ for the primal or dual setting, respectively :
\begin{multline}
	(P): \phi_v(\bm{x}_{v,i})
	= \bm{U}_{v}
	\big( \bm{\Gamma} - \bm{U}_{v}^\top \bm{U}_{v}
	\big)^{-1}\\
	\sum_{\substack{w=1 \\ w \ne v}}^V\bm{U}_{w}^{\top}\phi_w(\bm{x}_{w,i}) ,      \label{eq:pred_primal}
\end{multline}
\begin{multline}
	(D): \bm{k}_v(\bm{x}_{v,i})
	= \bm{K}_{v} \bm{H}
	\big( \bm{\Gamma} - \bm{H}\tran \bm{K}_{v} \bm{H}	\big)^{-1}\\
	\big(\bm{H}\tran\sum_{\substack{w=1 \\ w \ne v}}^V\bm{k}_v(\bm{x}_{v,i})\big), \label{eq:pred_dual}
\end{multline}
where $\bm{k}_v(\bm{x}_{v,i})=[k_v(\bm{x}_{v,1},\bm{x}_{v,i}),
\ldots,k_v(\bm{x}_{v,n},\bm{x}_{v,i})]\tran$. This can be used in various
\textit{pre-image methods} to impute a missing view. 

\noindent \textbf{Pre-image problem.}
Except some particular cases (e.g., with a linear kernel), one does not know the inverse of the feature maps $\psi_v(\cdot)\equiv\phi_v^{-1}(\cdot):\mathbb{R}^{d_{f,v}}\mapsto\mathbb{R}^{d_v}$. Furthermore, when using a kernel function, even the feature map $\phi_v(\cdot)$ is not known.
This gives rise to the pre-image problem: given a point $\Psi
	\in \mathcal{H}_v$, find $ \bm{x}^\star_v \in \mathcal{X}_v$ such that $\Psi =
	\phi(\bm{x}^\star_v)$.
This problem is known to be
ill-posed~\citep{mika_kernel_nodate} as the pre-image might not exist
\citep{Scholkopf2001} or, if it exists, it may not be unique.

\paragraph{Parametric pre-image map.}
One strategy to find the approximate pre-image $\bm{x}^\star_v$ is to formulate an optimization problem:
	$\arg\!\min_{\bm{x}^\star_v} \Vert \Psi-\phi(\bm{x}^\star_v)\Vert^{2}.$
It is possible to train a parametric feature map separately or one can jointly train a parametric pre-image map $\psi(\cdot;\bm{\eta})$ with the model. In this case, Algorithm \ref{alg:training_stiefel_P} or \ref{alg:training_stiefel_D} should be used and the objective function should be augmented with a reconstruction loss term  \cite{strkm, GENRKM}.
Note that besides this, various other techniques exist to approximately solve this
problem~\citep{
	kwok_pre-image_2004-2,
	weston_learning_2004,honeine_preimage_2011-1,joachim, bui_projection-free_2019}.
\begin{figure}[t]
	\centering
	\includegraphics[width=\linewidth]{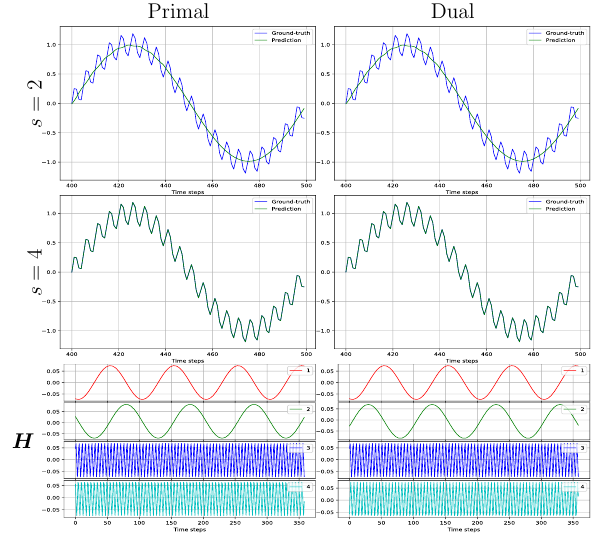}
	\caption{Forecasts (first two rows) and latent components (last row) obtained
		from the primal and dual models, trained by eigenvalue decomposition, with
		different number of components. Dataset: Sum of sine waves, $p=40$ and
		linear kernels on both views. Principal components ($\bm{H}\in
			\mathbb{R}^{n\times 4}$) obtained from training the primal (\cref{eq:SC2})
		and dual models (\cref{eq:kpca}).
		These components are used progressively to obtain forecasts.}
	\label{fig:sum_sine_svd}
\end{figure}
\begin{figure}[t]
	\centering
	\includegraphics[width=\linewidth]{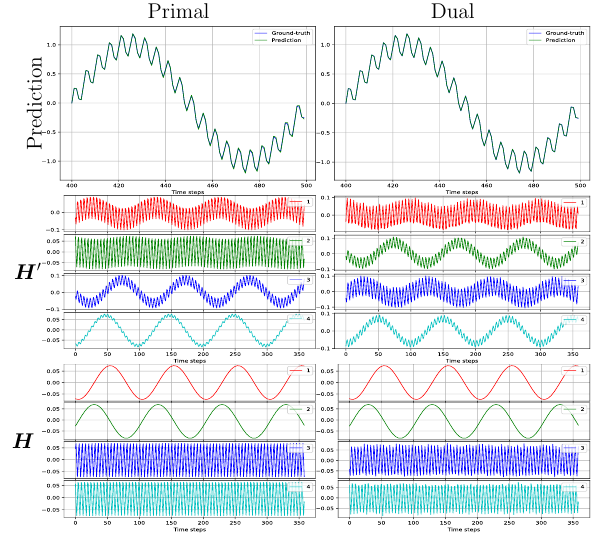}
	\caption{
    Forecasts (first row) and latent components (last two rows) obtained from
    the Stiefel training of primal and dual models on Sum of Sine waves dataset.
    See \cref{cor:Primal_rotations,corr:kpca_dual} for obtaining $\bm{H}^\prime$
    and $\bm{H}$.
	}
	\label{fig:sum_sine_stiefel}
\end{figure}
\begin{figure}[h!]
	\centering
	\includegraphics[width=\linewidth]{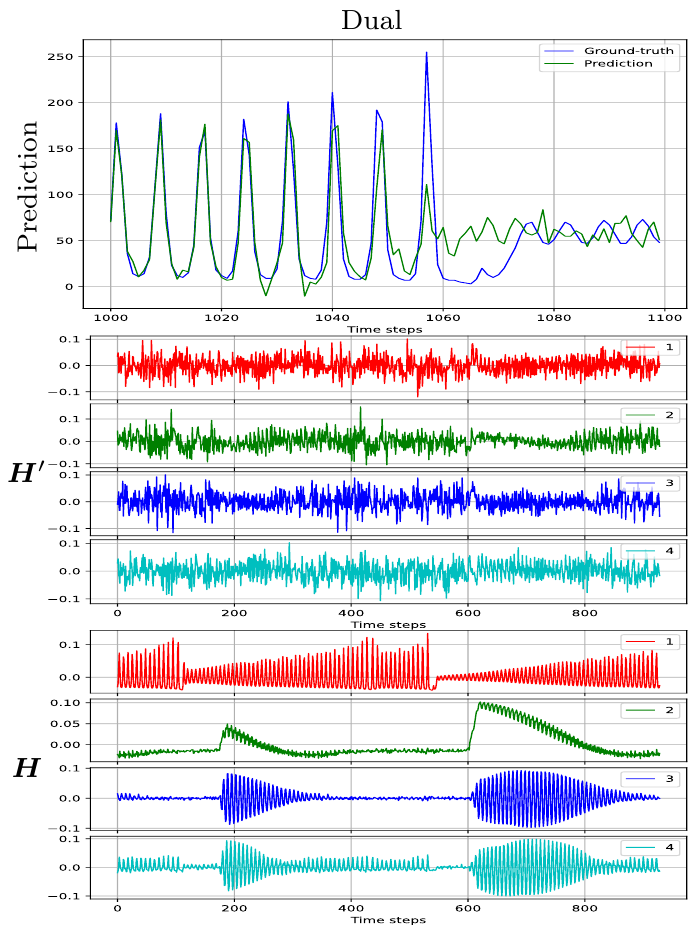}
	\caption{
		Forecasts ($1^{\text{st}}$ row) and latent components (last two rows) obtained from the Stiefel training of the dual model on SantaFe dataset. See \cref{cor:Primal_rotations,corr:kpca_dual} for obtaining $\bm{H}^\prime$ and $\bm{H}$.
	}
	\label{fig:santafe_st_dual}
\end{figure}

\begin{figure}[h]
	\centering
  \includegraphics[width=\linewidth]{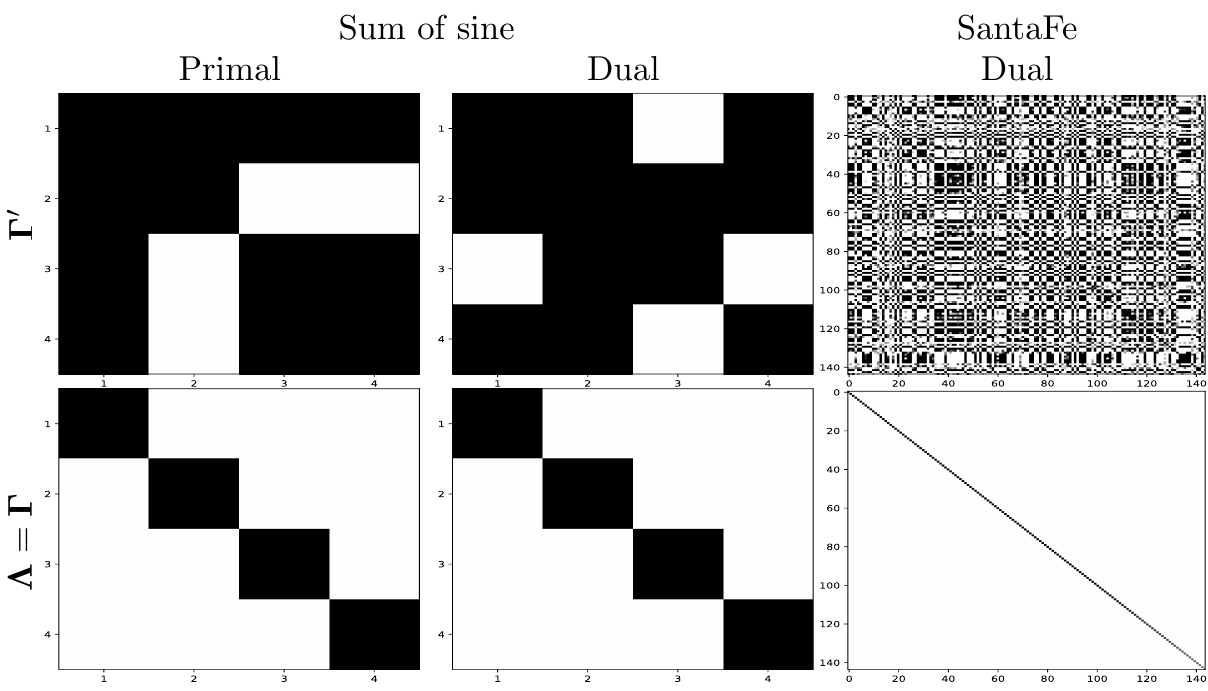}
	\caption{Illustration of the obtained $\bm{\Gamma}^{\prime}$ matrix of the models
		trained with the Stiefel optimizer, which contains many nonzero elements
		outside its diagonal. Their transformations $\bm{\Gamma}=\bm{\Gamma}^{\prime}\bm{O}$ equal
		the diagonal eigenvalue matrix $\bm{\Lambda}$ obtained by the eigendecomposition
		based algorithms (see \cref{cor:Primal_rotations,corr:kpca_dual}).}
	\label{fig:spectrum}
\end{figure}

\section{Experiments}
In this section, we perform experiments on publicly available standard datasets to validate our proposed framework.
In particular, we validate \cref{alg:training_eig_D,alg:training_eig_P,alg:training_stiefel_D,alg:training_stiefel_P} and the duality in the framework by analyzing the latent components $\bm{H}$ and the predictions $\bm{y}$. The code used to reproduce the experiments is provided on Github \footnote{\href{https://github.com/sonnyachten/dMVRKM}{https://github.com/sonnyachten/dMVRKM}}.

For simplicity and clarity, we consider two views ($V=2$), where one view is treated as the input source ($\mathcal{X}$) and the other view as the target ($\mathcal{Y}$) in the time series forecasting problem.
We use this 2-view model as the static regression problem and build a Non-linear Autoregressive model $\hat{{x}}_{\ell+1} = f( {x}_\ell , {x}_{\ell-1} , \ldots , {x}_{\ell-p} )$ with lag parameter $p\in \mathbb{N}$ and $\hat{{y}}$ is the estimated output.
Given a time-series of length $T$: $\{{x}_\ell\}_{\ell=1}^{T}$, we construct a training set with inputs $\bm{x}_{\ell}=x_{\ell:\ell-p}$ and corresponding outputs ${y}_{\ell}= {x}_{\ell+1}$, with $\ell$ ranging from $p+1$ to $T-1$. 
During the prediction phase, the output from the previous time-step is used as input in the current step. This is used \emph{iteratively} to forecast multiple steps ahead in future.

\noindent \textbf{Sum of Sinusoids.}
We begin with a toy-experiment to better understand different components of the model. Consider this synthetic time-series of the sum of two sine waves:
$x_\ell = \sum_{j=1}^{2} \alpha_j \sin(2
	\pi \omega_j \ell)$, $ \alpha_j \in \{ 1, 0.2 \}$, $\omega_j \in
	\{ 100, 2000 \}$.
We use linear kernels on both the views: $k_{{x}}(\cdot, \cdot)$ and $k_{{y}}(\cdot, \cdot)$.
Since linear kernels are used, feature maps and hence the pre-image maps are identity functions. In this experiment, we test whether the embeddings in the latent space (also called as principal components) and  the predictions are equivalent across all the \cref{alg:training_eig_D,alg:training_eig_P,alg:training_stiefel_D,alg:training_stiefel_P}.

\Cref{fig:sum_sine_svd} shows the forecasts and latent components when the primal and dual models are trained with eigenvalue decomposition. It also shows the ablation study of using the different number of components $s$ for prediction.  The figure clearly shows that each component progressively captures more variance in the dataset.
\begin{wraptable}{R}{3.4cm}
	\vspace*{-0.5cm}
	\caption{Experiment matrix for Sum of Sinusoids dataset}
 \label{tab:sine_exp_matrix}
	\begin{tabular}{c|c|c}
		      & Primal     & Dual       \\ \midrule
		Eig.  & \checkmark & \checkmark \\ \midrule
		$\St$ & \checkmark & \checkmark \\ \bottomrule
	\end{tabular}
	\vspace*{-0.5cm}
\end{wraptable}
Further, the forecasts from both primal and dual models are identical. Another interesting observation is that the latent components are also equivalent up to a \emph{sign-flip}. This corroborates the observation that the eigenvectors are unique up to a sign-flip.

\noindent \textbf{Significance of rescaling the primal variables.} We repeated the above experiment in primal setting, while omitting step 6 in \cref{alg:training_eig_P}. \Cref{fig:sum_sine_unscaled_U} in \cref{appendix_datasets} shows that the equivalence with the dual setting is only achieved up to a scaling for the dual components. Moreover, the figure also shows that omitting the rescaling has a detrimental effect on the predictions. This effect was not observed in \citet{strkm} because they trained a parametric pre-image map, combined with a reconstruction loss. Because of this, the rescaling was implicitly embedded in the pre-image map instead of the primal components themselves.

Next, we turn our attention to the Stiefel training of primal and dual models. See \cref{fig:sum_sine_stiefel} for the forecasts and the latent components as obtained by the Stiefel training $\bm{H}^\prime$ and rotated ones $\bm{H}$. 
\Cref{fig:spectrum} shows an illustration of the obtained hyperparameter matrices in both cases.
Note that such rotational transformations can always be obtained and are computationally easy to compute ($\mathcal{O}(s^3)$). The figure shows the equivalence between both the models when trained with optimization on Stiefel manifold or when trained with eigenvalue decomposition in \cref{fig:sum_sine_svd} (ref. \cref{tab:sine_exp_matrix}). 

\noindent \textbf{SantaFe.}
We repeat the above experiments on a real-world chaotic time-series dataset: SantaFe~\citep{santafe}.
Since the goal is to compare primal and dual models, the feature maps has to be explicitly known.
Therefore, we use the Random Fourier Features ($d_{f,x}=5000$) as an approximation of the Gaussian kernel on $k_x$ and linear kernel on $k_y$.
Since this linear kernel has identity feature map, no pre-image method is needed for predictions.
\Cref{fig:santafe_svd} in \cref{appendix_datasets} shows the latent components and the predictions for the SantaFe dataset when the primal and dual models are trained with eigenvalue decomposition.
We can see that the latent components and predictions are consistent with each other. We note that the goal of this experiment is not to have the best predictive performance, but to validate the duality in our model through different algorithms. Nevertheless, the predictive performance could be improved with further hyperparameter tuning.

\begin{wraptable}{r}{3.4cm}
	\vspace*{-0.5cm}
	\caption{Experiment matrix for SantaFe dataset}
 \label{tab:santafe_exp_matrix}
	\begin{tabular}{c|c|c}
		      & Primal     & Dual       \\ \midrule
		Eig.  & \checkmark & \checkmark \\ \midrule
		$\St$ & ---        & \checkmark \\ \bottomrule
	\end{tabular}
	\vspace*{-0.5cm}
\end{wraptable}
In the case of Stiefel training, by following the flowchart in~\cref{fig:schematic_flowchart}, we deduce that the \cref{alg:training_stiefel_D} is computationally efficient in this case.
This is because of the comparatively smaller parameter matrix given by the dual formulation.
Kernel matrix $\bm{K}\in \mathbb{R}^{n \times n}$, where $n=1000$ leads to a parameter matrix $\bm{H}\in \St(1000, s)$.
On the other hand, the covariance matrix $\bm{C}\in \mathbb{R}^{5001 \times 5001}$ is quite large due to the
very high-dimensional feature space ($d_{f,x}=5000$) given by the Random Fourier Features. This in turn, creates about 5 times larges parameter matrix for Stiefel optimization $\bm{U}\in \St(5001, s)$. Therefore, cleverly using the duality helps us to obtain a computational speed gain of 5x in this case. \Cref{fig:santafe_st_dual} shows the forecasts and latent components $\bm{H}^\prime$ and $\bm{H}$. By comparing it with \cref{fig:santafe_svd}, we conclude that the duality holds not just in theory but also in practice (ref. \cref{tab:santafe_exp_matrix}).

Lastly, the role of Stiefel training  becomes critical when training parametric feature/pre-image maps ($\phi_{v}(\cdot ; \bm{\theta}), \psi_{v}(\cdot ; \bm{\eta})$) such as deep neural networks where the parameter space is quite large. In such  cases, it becomes computationally infeasible to perform matrix decomposition in every mini-batch of the training loop.

Given that previous studies \cite{TONIN2021661, GENRKM, strkm, pandey_multi-view_2023} have already benchmarked and demonstrated the soundness and effectiveness of individual algorithms, the primary objective of this paper is to enhance our understanding of the relationships between these methods, by assessing the equivalence of their core models. Therefore, we did not focus on performances and even judged that using models with sub-optimal performance are better to (visually) assess the equivalence.

\section{Conclusion}
We explored and highlighted duality in Multi-view Kernel PCA. Further, we provided a flowchart of four algorithms to train the models depending on whether the feature maps are known or not, whether they are parametric or not and the number of data-points versus the dimensionality of the feature spaces. By rescaling primal variables, we showed the equivalence of the primal and dual setting. We show how to rotate the solution obtained by Stiefel optimization based algorithms to match the components of the eigendecompositions. Lastly, we validated our model and algorithms on publicly available standard datasets and performed ablation studies where we empirically verified the equivalence.
Future work includes investigating equivalence for generative modeling. Furthermore, how to enable mini-batch training in the dual formulation remains an open question.

\section*{Acknowledgements}
European Research Council under the European
Union's Horizon 2020 research and innovation programme: ERC Advanced Grants
agreements E-DUALITY(No 787960) and Back to the Roots (No 885682). This paper
reflects only the authors' views and the Union is not liable for any use that
may be made of the contained information. Research Council KUL: Optimization
frameworks for deep kernel machines C14/18/068; Research Fund (projects
C16/15/059, C3/19/053, C24/18/022, C3/20/117, C3I-21-00316); Industrial Research
Fund (Fellowships 13-0260, IOFm/16/004, IOFm/20/002) and several Leuven Research
and Development bilateral industrial projects. Flemish Government Agencies: FWO:
projects: GOA4917N (Deep Restricted Kernel Machines: Methods and Foundations),
PhD/Postdoc grant, EOS Project no G0F6718N (SeLMA), SBO project S005319N,
Infrastructure project I013218N, TBM Project T001919N, PhD Grant (SB/1SA1319N);
EWI: the Flanders AI Research Program; VLAIO: CSBO (HBC.2021.0076) Baekeland PhD
(HBC.20192204). This research received funding from the Flemish Government (AI
Research Program). Other funding: Foundation `Kom op tegen Kanker', CM
(Christelijke Mutualiteit). Sonny Achten, Arun Pandey, Hannes De Meulemeester, Bart De Moor and Johan Suykens are also affiliated with Leuven.AI - KU Leuven institute for AI, B-3000, Leuven, Belgium.

\FloatBarrier
\bibliography{example_paper}
\bibliographystyle{icml2023}

\newpage

\onecolumn
\appendix

\section{List of symbols}\label{app:list_of_symbols}

\begin{table}[h]
	\begin{center}
		\begin{small}
			\begin{tabular}{lll}
				\toprule
				Symbol                 & Space                                                         & Description                                                                                               \\
				\midrule
				$\matr{1}_n$           & $\{1\}^{n}$                                                   & $n$-dimensional vector of all ones                                                                        \\
				$\boldsymbol{\Gamma}$  & $\mathbb{R}_{\succ 0}^{s \times s}$                           & symmetric positive definite hyperparameter matrix                                                         \\
				$\delta_{ij}$          & $\{0,1\}$                                                     & Kronecker delta                                                                                           \\
				$\boldsymbol{\eta}$    & $\mathbb{R}^{t}$                                              & parametervector of a parametric pre-image map with $t$ parameters                                         \\
				$\boldsymbol{\theta}$  & $\mathbb{R}^{t}$                                              & parametervector of a parametric feature map with $t$ parameters                                           \\
				$\boldsymbol{\Lambda}$ & $\mathbb{R}_{\succ 0}^{s \times s}$                           & diagonal matrix containing eigenvalues                                                                    \\
				$\phi_v(\cdot)$        & $\mathbb{R}^{d} \mapsto \mathbb{R}^{d_{f,v}}$                 & feature map for view $v$, transforming an input from a $d$-dimensional space to a $d_f$-dimensional space \\
				$\Phi_v$               & $\mathbb{R}^{n\times d_{f,v}}$                                & matrix containing all feature maps as row vectors                                                         \\

				$\psi_v(\cdot)$        & $\mathbb{R}^{d_{f,v}}\mapsto\mathbb{R}^{d_v}$                 & pre-image map of view $v$ $\psi_v(\cdot)\equiv\phi_v^{-1}(\cdot)$
				\\
				$\matr{C}$             & $\mathbb{R}^{d_f \times d_f}$                                 & cross-covariance matrix (see \eqref{eq:C_mat})                                                             \\
				$\matr{C}_{vw}$        & $\mathbb{R}^{d_{f,v} \times d_{f,w}}$                         & cross-covariance matrix of views $v$ and $w$                                                               \\
				$d_f$                  & $\mathbb{N}$                                                  & sum of feature space dimensionalities of all views
				\\
				$d_{f,v}$              & $\mathbb{N}$                                                  & feature space dimensionality of view $v$
				\\
				$d_v$                  & $\mathbb{N}$                                                  & input space dimensionality of view $v$
				\\
				$\matr{e}_{v,i}$       & $\mathbb{R}^{s}$                                              & score variable of datapoint $i$ for view $v$                                                              \\
				$\matr{h}_i$           & $\mathbb{R}^{s}$                                              & dual representation of datapoint $i$                                                                      \\
				$\matr{H}$             & $\mathbb{R}^{n\times s}$                                      & matrix containing all dual representations as row vectors                                                 \\
				$\mathbb{I}_n$         & $\{0,1\}^{n\times n}$                                         & $n$ by $n$ identity matrix                                                                                \\
				$k_v(\cdot,\cdot)$     & $\mathbb{R}^{d_v} \times \mathbb{R}^{d_v} \mapsto \mathbb{R}$ & a positive definite kernel function                                                                       \\
				$\matr{K}$             & $\mathbb{R}^{n\times n}$                                      & summation of all $\matr{K}_v$                                                                             \\
				$\matr{K}_v$           & $\mathbb{R}^{n\times n}$                                      & kernel matrix

                    $K_{v,ij}=k(\matr{x}_{v,i},\matr{x}_{v,j})$, or the 
				Gram matrix $K_{v,ij}=\langle \phi_v(\matr{x}_{v,i}),\phi_v(\matr{x}_{v,j})\rangle_{\mathcal{H}}$ for view $v$                                                                                                                \\
    $\ell$                  & $\mathbb{N}$ & timestep index of timeseries \\
				$\matr{M}_c$           & $\mathbb{R}^{n \times n}$                                     & centering matrix, used to center the kernel matrix                                                        \\
				$n$                    & $\mathbb{N}$                                                  & number of datapoints                                                                                      \\
				$p$                    & $\mathbb{N}$                                                  & number of lags                                                                                            \\
				$s$                    & $\mathbb{R}$                                                  & dimensionality of the score variables and dual representations, i.e., number of components                \\
				$\matr{U}$             & $\mathbb{R}^{d_f\times s}$                                    & concatenation of all $\matr{U}_v$                                                                         \\
				$\matr{U}_v$           & $\mathbb{R}^{d_{f,v}\times s}$                                    & linear transformation matrix for view $v$                                                                 \\
				$V$                    & $\mathbb{N}$                                                  & number of views (data modalities)                                                                         \\
				$\matr{x}_{v,i}$       & $\mathbb{R}^{d_v}$                                            & the feature vector of datapoint $i$ for view $v$                                                          \\
				${y}_i$                & $\mathbb{R}$                                                  & target variable of datapoint $i$                                                                          \\
				$\matr{Z}$             & $\mathbb{R}^{s \times s}$                                     & a matrix containing Lagrange multipliers                                                                  \\
				\bottomrule
			\end{tabular}
		\end{small}
	\end{center}
	\vskip -0.1in
\end{table}
\section{Proofs}\label{app:proofs}

\subsection{Proof of Lemma \ref{lemma:primal_equivalence}}

\begin{proof}
	By eliminating dual variables $\bm{h}_i$ from \eqref{eq:SC1} using \eqref{eq:SC2}, one obtains:
	\begin{equation}
		\sum_{w=1}^V\bm{C}_{vw}\bm{U}_w=\bm{U}_v\bm{\Gamma}, \quad \forall v, \label{eq:SC2inSC1}
	\end{equation}
	and by substituting \eqref{eq:SC2} in \eqref{eq:SC3}:
	\begin{equation}
		\phi_v(\bm{x}_{v,i}) =
		\bm{U}_{v} \bm{\Gamma}^{-1}\sum_{w=1}^V\bm{U}_w\tran\phi_w(\bm{x}_{w,i}). \label{eq:SC2inSC3}
	\end{equation}

	Next, the Lagrangian of \eqref{eq:primal_optimization_problem} is:
	\begin{equation*}
		\mathcal{L}(\tilde{\bm{U}},\bm{Z})=J_{\text{pr}}+\frac{1}{2}\Tr\left(\bm{Z}\tran\left(-\mathbb{I}_s+\sum_{v=1}^V\tilde{\bm{U}}_v\tran\tilde{\bm{U}}_v\right)\right),
	\end{equation*}
	and the Karush-Kuhn-Tucker conditions are:

	\begin{empheq}[left=\empheqlbrace]{align}
		\dfrac{\partial \mathcal{L}}{\partial \tilde{\bm{U}}_v} = 0 \Rightarrow & \ \sum_{w=1}^V\bm{C}_{vw}\tilde{\bm{U}}_w = \tilde{\bm{U}}_v\Tilde{\bm{Z}}, & \forall v, \label{eq:primal_KKT_1}\\
		\dfrac{\partial \mathcal{L}}{\partial \phi_v(\bm{x}_{v,i})}= 0 \Rightarrow & \ \phi_v(\bm{x}_{v,i}) =
		\tilde{\bm{U}}_{v}\sum_{w=1}^V\tilde{\bm{U}}_w\tran\phi_w(\bm{x}_{w,i}), & \forall v,i\label{eq:primal_KKT_2}\\
		\dfrac{\partial \mathcal{L}}{\partial \bm{Z}} = 0 \Rightarrow & \ \sum_{v=1}^V\tilde{\bm{U}}_v\tran\tilde{\bm{U}}_v=\mathbb{I}_s, & \forall v.\label{eq:primal_KKT_3}
	\end{empheq}

	By replacing $\tilde{\bm{U}}_v$ in the KKT conditions with $\bm{U}_v\tilde{\bm{Z}}^{-1/2}$, and by choosing $\bm{\Gamma}=\tilde{\bm{Z}}$, we obtain \eqref{eq:SC2inSC1} and \eqref{eq:SC2inSC3} from \eqref{eq:primal_KKT_1} and \eqref{eq:primal_KKT_2} respectively, which proves the Lemma.
\end{proof}

\subsection{Proof of Proposition \ref{proposition:primal_span}}\label{app:proof_prop_span}

We use the proof as given by \citet[Appendix]{achten2023semisupervised}, and rewrite it for the variables $\bm{A}=[\bm{a}_1,\dots,\bm{a}_m]\tran$ and $\bm{B}$.

\begin{proof}
	The Lagrangian of \eqref{eq:proposition_objective} is:
	\begin{equation*}
		\mathcal{L}(\matr{A},\matr{Z})=-\frac{1}{2}\text{Tr}(\matr{A}\tran \matr{B} \matr{A})+\frac{1}{2}\text{Tr}(\bm{B})+\frac{1}{2}\text{Tr}(\matr{Z}\tran(\matr{A}\tran\matr{A}-\mathbb{I}_n)),
	\end{equation*}
	and the Karush-Kuhn-Tucker conditions are:
	\begin{equation}\label{eq:proof-KKT}
		\left\{ \begin{array}{l}
			\frac{\partial \mathcal{L}}{\partial \matr{A}} = -\matr{B}\matr{A}+\matr{A}(\matr{Z}+\matr{Z}\tran)/2=0 \\\\
			\frac{\partial \mathcal{L}}{\partial \matr{Z}} = \matr{A}\tran\matr{A}-\mathbb{I}_n = \bm{0}_n.\end{array}
		\right.
	\end{equation}

	Let us define the columns of $\matr{A}$ as $\matr{g}_i = [(\matr{a}_1)_i, \dots,(\matr{a}_m)_i]\tran$, and $\matr{G} = [\matr{g}_1, \dots,\matr{g}_n] = \matr{A}$. By also defining $\tilde{\bm{Z}}=(\matr{Z}+\matr{Z}\tran)/2$, we can then rewrite \eqref{eq:proof-KKT} in vector notation:
	\begin{equation}\label{app:eq:GCKM_stat3}
		\left\{ \begin{array}{l}
			\frac{\partial \mathcal{L}}{\partial \matr{g}_i} = -\frac{1}{\eta}\matr{B}\matr{g}_i +\sum_{j=1}^n\tilde{Z}_{ij}\matr{g}_j=\matr{0}_m \quad \forall i= 1\dots n \\\\
			\frac{\partial \mathcal{L}}{\partial \tilde{Z}_{ij}} = \matr{g}_i\tran\matr{g}_j-\delta_{ij}=0 \quad \forall i,j = 1 \dots n\end{array}
		\right.
	\end{equation}
	where $\delta_{ij}$ is the Kronecker delta, and derive the second order derivatives for the optimization parameters:
	\begin{equation}
		\nabla^2_{\matr{g}_i\matr{g}_j}\mathcal{L}=-\frac{\delta_{ij}}{\eta}\matr{B} + \tilde{Z}_{ij}\mathbb{I}_m.
	\end{equation}
	By defining $\matr{D} = [\matr{d}_1, \dots,\matr{d}_n]$, the second order necessary conditions can be formulated as:
	\begin{gather}
		\sum_{i=1}^n\sum_{j=1}^n\matr{d}_i\tran(-\frac{\delta_{ij}}{\eta}\matr{B} + \tilde{Z}_{ij}\mathbb{I}_m)\matr{d}_j=-\frac{1}{\eta}\sum_{i=1}^n\matr{d}_i\tran\matr{B}\matr{d}_i+\sum_{i=1}^n\sum_{j=1}^n\tilde{Z}_{ij}\matr{d}_i\tran\matr{d}_j\geq0 \nonumber\\
		\forall \matr{D} \in C(\matr{G}^*)\label{app:eq:SONC},
	\end{gather}
	where $C(\matr{G}^*)=\{\matr{D}\in\mathbb{R}^{m\times n} \ | \ \matr{D}\tran\matr{G}^*=\matr{0}_n\}$ is the critical cone at $\matr{G}^*$. For the critical cone, it can be deduced that the following properties hold:
	\begin{flalign*}
		\matr{d}_i\tran\matr{g}^*_j+\matr{d}_j\tran\matr{g}^*_i=0 & \quad \forall i,j=1,\dots, n              \\
		\matr{d}_i\tran\matr{g}^*_i=0                             & \quad \forall i=1,\dots, n                \\
		\matr{d}_i\tran\matr{d}_j=0                               & \quad \forall i,j=1,\dots, n\quad i\ne j.
	\end{flalign*}
	Without loss of generality, we further assume $\lVert\matr{d}_i\rVert=1$. From \eqref{app:eq:GCKM_stat3}, one can derive $\tilde{Z}_{ij}=\frac{1}{\eta}\matr{g}^{*\top}_j\matr{B}\matr{g}^*_i$. By substituting this and $\matr{d}_i\tran\matr{d}_j=0$ in \eqref{app:eq:SONC}, the second order necessary condition becomes:
	\begin{equation*}
		\sum_{i=1}^n\frac{1}{\eta}\matr{g}^{*\top}_i\matr{B}\matr{g}^*_i \geq \sum_{i=1}^n\frac{1}{\eta}\matr{d}_i\tran\matr{B}\matr{d}_i,
	\end{equation*}
	or after reworking this algebraically:
	\begin{equation}\label{app:eq:reworkedSONC}
		\text{Tr}(\matr{G}^{*\top}\matr{V}\boldsymbol{\Lambda}\matr{V}\tran\matr{G}^*) \geq \text{Tr}(\matr{D}\tran\matr{V}\boldsymbol{\Lambda}\matr{V}\tran\matr{D}),
	\end{equation}
	with $\matr{V}=[\matr{v}_1,\dots,\matr{v}_m]$ the eigenvectors of $\matr{B}$ with corresponding eigenvalues $\boldsymbol{\Lambda}=\text{diag}(\lambda_1,\dots,\lambda_m)$. For the case where $\text{span}(\matr{g}^*_1, \dots, \matr{g}^*_n)=\text{span}(\matr{v}_1, \dots, \matr{v}_n)$, the left hand side is maximal:
	\begin{equation*}
		\sum_{i=1}^n\lambda_i = \text{Tr}(\matr{G}^{*\top}\matr{V}\boldsymbol{\Lambda}\matr{V}\tran\matr{G}^*) \geq \text{Tr}(\matr{D}\tran\matr{V}\boldsymbol{\Lambda}\matr{V}\tran\matr{D}).
	\end{equation*}
	The right hand side becomes maximal when $\text{span}(\matr{d}_1, \dots, \matr{d}_n)=\text{span}(\matr{v}_1, \dots, \matr{v}_n)$. In this case, there exists an orthonormal transformation matrix $\bm{O}$ such that $\matr{G}^*=\matr{D}\matr{O}$:
	\begin{equation*}
		\sum_{i=1}^n\lambda_i = \text{Tr}(\matr{G}^{*\top}\matr{V}\boldsymbol{\Lambda}\matr{V}\tran\matr{G}^*) = \text{Tr}(\matr{O}\tran\matr{D}^{T}\matr{V}\boldsymbol{\Lambda}\matr{V}\tran\matr{D}\matr{O})=\text{Tr}(\matr{D}\tran\matr{V}\boldsymbol{\Lambda}\matr{V}\tran\matr{D}).
	\end{equation*}

	We verified that the second order necessary conditions are satisfied for in the case $\text{span}(\matr{g}^*_1, \dots, \matr{g}^*_n) = \text{span}(\matr{v}_1, \dots, \matr{v}_n)$.
	Let us now proceed by assuming that $\text{span}(\matr{g}^*_1, \dots, \matr{g}^*_n) \ne \text{span}(\matr{v}_1, \dots, \matr{v}_n)$. In this case there exists a matrix $\bm{D}$ such that \eqref{app:eq:reworkedSONC} becomes:
	\begin{equation*}
		\text{Tr}(\matr{G}^{*\top}\matr{V}\boldsymbol{\Lambda}\matr{V}\tran\matr{G}^*) < \text{Tr}(\matr{D}\tran\matr{V}\boldsymbol{\Lambda}\matr{V}\tran\matr{D})\leq\sum_{i=1}^n\lambda_i.
	\end{equation*}

	We thus established that the second order necessary conditions are satisfied if and only if $\text{span}(\matr{g}^*_1, \dots, \matr{g}^*_n)=\text{span}(\matr{v}_1, \dots, \matr{v}_n)$. Generally, the second order condition is not sufficient. However, as the feasible set $\{\matr{A}\in\mathbb{R}^{m\times n} \ | \ \matr{A}\tran\matr{A}-\mathbb{I}_n = \matr{0}_n \}$ is compact, and the objective function $-\frac{1}{2}\text{Tr}(\matr{A}\tran \matr{B} \matr{A}) + \text{Tr}(\matr{B})$ is concave, \eqref{eq:proposition_objective} is guaranteed to have a global minimizer \cite{Boyd:2004:CO:993483}. Therefore, any $\matr{A}^*$ such that $\text{range}(\matr{A}^*)=span(\matr{v}_1 \dots \matr{v}_n)$ is a global minimizer.

\end{proof}

\subsection{Proof of Lemma \ref{lemma:dual_equivalence}}
\begin{proof}
	By eliminating primal variables $\bm{U}_v$ from \eqref{eq:SC2} using \eqref{eq:SC1}, one obtains:
	\begin{equation}
		\bm{K}\bm{H} = \bm{H} \bm{\Gamma}, \label{eq: dual decomposition}
	\end{equation}
	where $\bm{K}=\sum_{v=1}^V\bm{K}_{v}$, and by substituting \eqref{eq:SC1} in \eqref{eq:SC3}:
	\begin{equation}
		\phi_v(\bm{x}_{v,i}) = \sum_{j=1}^n\phi_v(\bm{x}_{v,j})
		\bm{h}_j\tran\bm{h}_{i}. \label{eq:SC1inSC3}
	\end{equation}

	Next, the Lagrangian of \eqref{eq:dual_optimization_problem} is:
	\begin{equation*}
		\mathcal{L}({\bm{H}},\bm{Z})=J_{\text{d}}+\frac{1}{2}\Tr\left(\bm{Z}\tran\left(-\mathbb{I}_s+{\bm{H}}\tran{\bm{H}}\right)\right),
	\end{equation*}
	and the Karush-Kuhn-Tucker conditions are:
	\begin{empheq}[left=\empheqlbrace]{align}
		\dfrac{\partial \mathcal{L}}{\partial {\bm{H}}} = 0 \Rightarrow \ &  [\sum_{v=1}^V\bm{K}_v]{\bm{H}} = {\bm{H}}\Tilde{\bm{Z}} \label{eq:dual_KKT_1}\\
		\dfrac{\partial \mathcal{L}}{\partial \phi_v(\bm{x}_{v,i})}= 0 \Rightarrow \ & \phi_v(\bm{x}_{v,i}) =  \sum_{j=1}^n\phi_v(\bm{x}_{v,j})
		\bm{h}_j\tran\bm{h}_{i} \label{eq:dual_KKT_2}\\
		\dfrac{\partial \mathcal{L}}{\partial \bm{Z}} = 0 \Rightarrow & \ {\bm{H}}\tran{\bm{H}}=\mathbb{I}_s. \label{eq:dual_KKT_3}
	\end{empheq}
	By choosing $\bm{\Gamma}=\tilde{\bm{Z}}$, we obtain \eqref{eq: dual decomposition} from \eqref{eq:dual_KKT_1}. Since also \eqref{eq:SC1inSC3} equals \eqref{eq:dual_KKT_2}, we have proven the Lemma.
\end{proof}


\section{Datasets and additional experiments \label{appendix_datasets}}

We refer to~\cref{Table:dataset}  for more  details  on  model  architectures, datasets and hyperparameters used in this paper. The PyTorch library (double precision) in Python was used. See~\cref{alg:training_eig_D,alg:training_eig_P,alg:training_stiefel_D,alg:training_stiefel_P} for training the MV-RKM model.

\begin{table}[ht]
	\caption{Datasets used for the experiments. $n_{\text{train}}$, $n_{\text{test}}$  is the number of training and test samples respectively, $d$ the time series dimension, $p$ is the lag, $s$ is the latent space dimension, RFF and RBF mean 'random Fourier features' and 'radial basis function' respectively.}
	\label{Table:dataset}
	\centering
	\begin{tabular}{lllll l l l l l}
		\toprule
		\textbf{Dataset} & $n_{\text{train}}$ & $n_{\text{test}}$ & $d$ & $p$ & $\phi_x$ & $k_x$  & $\phi_y$ & $k_y$  & $s$ \\ \midrule
		Sum of Sine      & 400                & 100               & 1   & 40  & Linear   & Linear & Linear   & Linear & 4   \\
		SantaFe          & 1000               & 100               & 1   & 70  & RFF      & RBF    & Linear   & Linear & 144 \\ \bottomrule
	\end{tabular}
\end{table}

\begin{figure}[ht]
	\centering
	\includegraphics[width=0.95\textwidth]{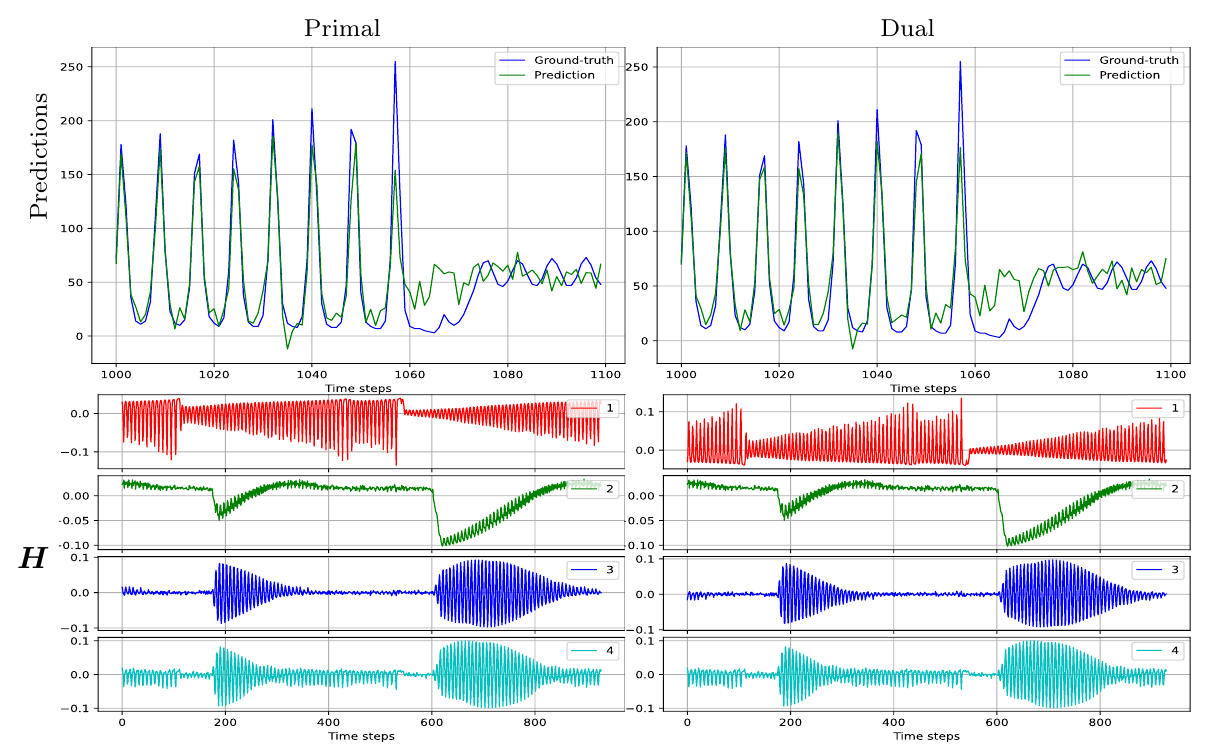}
	\caption{Forecasts ($1^{\text{st}}$ row) and top-4 latent components (last rows) obtained from the primal and dual model trained by eigenvalue decomposition. Dataset: SantaFe,  $k_x$ is Gaussian kernel ($\sigma = 2.1856$) approximated by Random Fourier Features ($d_{f,x}=5000$), $k_y$ is linear kernel, $p=70$,  $s=144$.
	}
	\label{fig:santafe_svd}
\end{figure}


\begin{figure}
	\centering
	\includegraphics[width=0.95\textwidth]{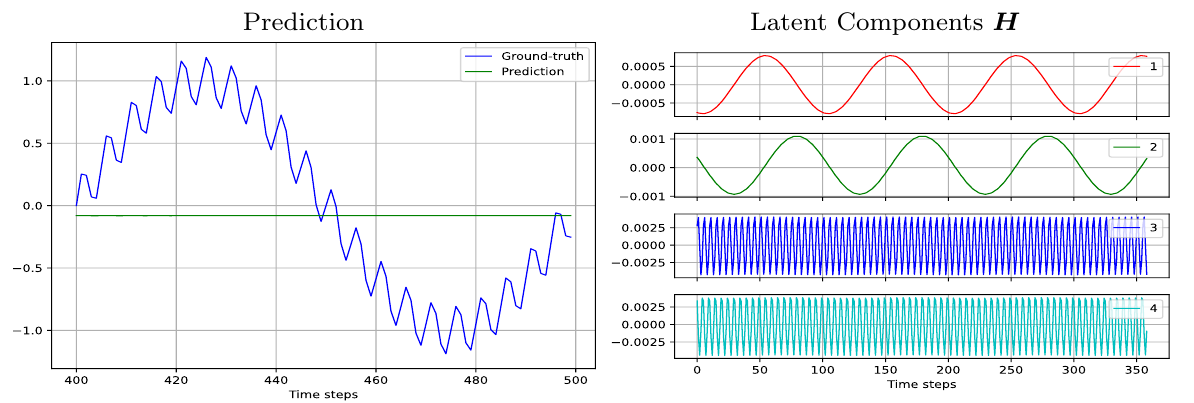}
	\caption{
		Forecasts and latent components obtained from the eigenvalue decomposition training of primal model on Sum of Sine waves dataset. This is done \emph{without} scaling $\bm{U}$ as proposed in step 6 of \cref{alg:training_eig_P} i.e. setting $\bm{U} = \tilde{\bm{U}}$. Notice that the latent components are equivalent as in \cref{fig:sum_sine_svd} but only upto a scaling (and sign-flip). However on the predictions,  this has a severely adverse effect and the model does not learn any variance in the dataset.
	}
	\label{fig:sum_sine_unscaled_U}
\end{figure}

\FloatBarrier

\end{document}